\documentclass[runningheads]{llncs}

\usepackage{eccv}

\usepackage{eccvabbrv}

\usepackage{graphicx}
\usepackage{booktabs}
\usepackage{multicol}
\usepackage{multirow}

\usepackage[accsupp]{axessibility}  

\usepackage[pagebackref,breaklinks,colorlinks]{hyperref}

\usepackage{orcidlink}

\begin{document}

\title{LoCo: Locally Constrained Training-Free Layout-to-Image Synthesis} 


\author{Peiang Zhao\inst{1} \and
Han Li\inst{1} \and
Ruiyang Jin\inst{1} \and
S. Kevin Zhou\inst{1,2}
}

\authorrunning{P. Zhao et al.}

\institute{
Center for Medical Imaging, Robotics, Analytic Computing \& Learning (MIRACLE), Suzhou Institute for Advanced Research, University of Science and Technology of China
\and
Key Lab of Intelligent Information Processing of Chinese Academy of Sciences (CAS), Institute of Computing Technology 
}

\maketitle

\begin{abstract}
Recent text-to-image diffusion models have reached an unprecedented level in generating high-quality images. 
However, their exclusive reliance on textual prompts often falls short in precise control of image compositions.
In this paper, we propose LoCo, a training-free approach for layout-to-image Synthesis that excels in producing high-quality images aligned with both textual prompts and layout instructions.  
Specifically, we introduce a Localized Attention Constraint (LAC), leveraging semantic affinity between pixels in self-attention maps to create precise representations of desired objects and effectively ensure the accurate placement of objects in designated regions.
We further propose a Padding Token Constraint (PTC) to leverage the semantic information embedded in previously neglected padding tokens, improving the consistency between object appearance and layout instructions.
LoCo seamlessly integrates into existing text-to-image and layout-to-image models, enhancing their performance in spatial control and addressing semantic failures observed in prior methods.
Extensive experiments showcase the superiority of our approach, surpassing existing state-of-the-art training-free layout-to-image methods both qualitatively and quantitatively across multiple benchmarks.
\end{abstract}

\section{Introduction}
\label{sec:intro}

Recently, text-to-image (T2I) diffusion models \cite{ramesh2022hierarchical,rombach2022high} have demonstrated unprecedented capacity for synthesizing high-quality images. 
Despite these accomplishments, these T2I models encounter a significant challenge: they depend solely on textual prompts for spatial composition control, which proves inadequate for various applications. 
For instance, in movie poster design, where multiple objects and attributes exhibit complex spatial relationships, dependence solely on position-related prompts for accurate object placement is inefficient and imprecise.
While texts can harness a rich repository of high-level concepts, they struggle to convey the fine-grained spatial composition of an image accurately.
Utilizing position-related prompts like "on the left" and "beneath" can only offer rudimentary spatial control, requiring users to sift through a pile of generated images to find satisfying results.
This challenge becomes more pronounced when the prompts becomes intricate or involves unusual scenes.

\begin{figure}[t]
\centering
\includegraphics[width=\linewidth]{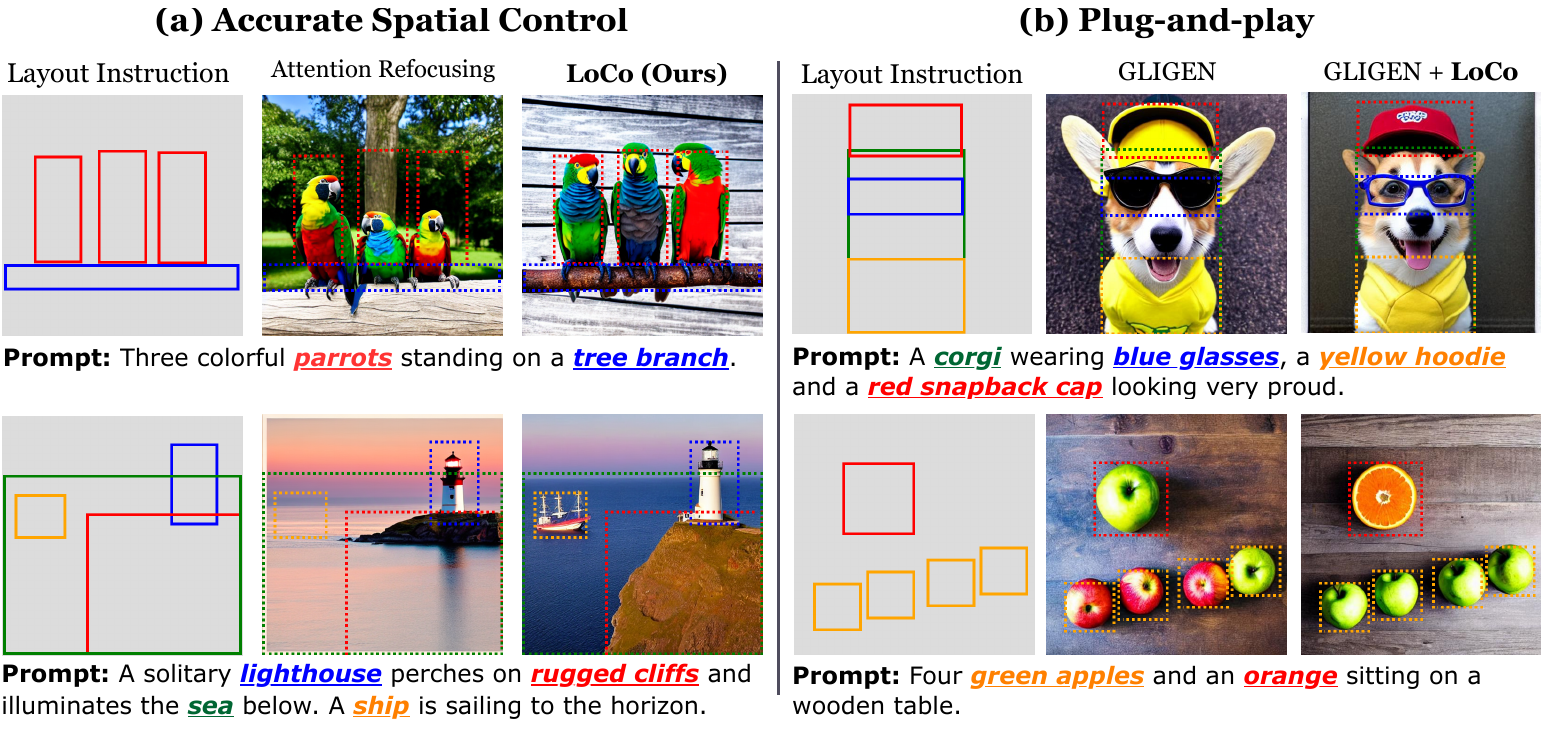}
\caption{
\textbf{(a) Accurate Spatial Control.} Existing training-free layout-to-image synthesis (LIS) approaches struggle to generate high-quality images that adhere to the given layout instructions. In contrast, LoCo is able to provide accurate spatial control. \textbf{(b) Plug-and-play.} LoCo can be integrated to fully-supervised LIS methods, \eg, GLIGEN \cite{li2023gligen}, serving as a plug-and-play booster to enhance their performance.}
\label{fig:padding_tokens}
\end{figure}
To address this challenge, researchers have explored layout-to-image synthesis (LIS) methods \cite{zhao2019image,sylvain2021object,sun2021learning,li2021image,gafni2022make,avrahami2023spatext,cheng2023layoutdiffuse,zeng2023scenecomposer,li2023gligen,zhang2023adding}. These methods allow users to specify the locations of objects with various forms of layout instructions, \eg, bounding boxes, semantic masks, or scribbles. Generally, these layout-to-image approaches can be categorized into two types: fully-supervised methods and training-free methods.

Fully-supervised layout-to-image methods have shown remarkable results, either by training new layout-to-image models \cite{zeng2023scenecomposer, wang2022semantic} or by enhancing existing T2I models with auxiliary modules \cite{li2023gligen, zhang2023adding, mou2023t2i, zhou2024migc, wang2024instancediffusion} to incorporate layout instructions.
Unfortunately, these approaches demand substantial amounts of paired layout-image training data, which is expensive and challenging to obtain. Additionally, both training and fine-tuning a model are computationally intensive. 

On the contrary, a noteworthy line of research \cite{xie2023boxdiff,chen2023training,phung2023grounded,couairon2023zero, xiao2023r} demonstrates that layout-to-image synthesis can be achieved in a training-free manner. 
Specifically, they guide the synthesis process by updating the latent feature based on cross-attention maps extracted at each timestep. 
However, since cross-attention maps predominantly capture prominent parts of the objects, they serve as coarse-grained and noisy representations of desired objects. Therefore, directly using cross-attention maps to guide the synthesis process only offers \textit{limited spatial controllability}. Specifically, the synthesized objects often deviate from their corresponding layout instructions, resulting in unsatisfactory outcomes.
Besides, these methods also suffer from \textit{semantic failures}, \eg, missing or fused objects, incorrect attribute binding, \etc. 

To address these issues, we introduce \textbf{LoCo}, short for Locally Constrained Diffusion, a novel training-free approach designed to enhance spatial controllability in layout-to-image and alleviate semantic failures faced by previous methods.
Specifically, we propose two novel constraints, the \textbf{Localized Attention Constraint ($\mathcal{L}_{LAC}$)} and the \textbf{Padding Tokens Constraint ($\mathcal{L}_{PTC}$)}, to guide the synthesis process based on attention maps.

$\mathcal{L}_{LAC}$ aims to ensure the accurate generation of desired objects.
Departing from prior approaches that depend solely on coarse-grained cross-attention maps for spatial control, we leverage Self-Attention Enhancement to attain precise representations of the desired objects. Thus, $\mathcal{L}_{LAC}$ offers more accurate spatial control, enhancing the alignment between cross-attention maps and layout instructions and rectifying semantic failures.
The $\mathcal{L}_{PTC}$ taps into previously overlooked semantic information carried by padding tokens, specifically start-of-text tokens (\texttt{[SoT]}) and end-of-text tokens (\texttt{[EoT]}) in textual embedding. These tokens hold significant associations with the layout of the synthesized image. 
By harnessing this information, $\mathcal{L}_{PTC}$ prevents closely located objects from extending beyond their designated boxes and enhances the consistency between object appearance and layout instructions.

We perform comprehensive experiments, comparing our method with various approaches in the training-free layout-to-image synthesis literature. Our results demonstrate state-of-the-art performances, showcasing improvements both quantitatively and qualitatively over prior approaches. Additionally, our method can be integrated into fully-supervised layout-to-image synthesis methods, serving as a plug-and-play booster, consistently enhancing their performance.

In summary, our contributions are as follows:

\begin{itemize}
    \item We introduce LoCo, a training-free method for layout-to-image synthesis that excels in producing high-quality images aligned with both textual prompts and spatial layouts.
    \item We present two novel constraints, $\mathcal{L}_{LAC}$ and $\mathcal{L}_{PTC}$. The former provides precise spatial control and improves the alignment between synthesized images and layout instructions. The latter leverages the semantic information embedded in previously neglected padding tokens, further enhancing the consistency between object appearance and layout instructions.
    \item We conduct comprehensive experiments, comparing our approach with existing methods in the layout-to-image synthesis literature. The results showcase that LoCo outperforms prior state-of-the-art approaches, considering both quantitative metrics and qualitative assessments.
\end{itemize}

\section{Related Work}
\subsection{Text-to-image Diffusion models}
Large-scale text-to-image (T2I) diffusion models have garnered substantial attention due to their remarkable performances. For instance, Ramesh \etal \cite{ramesh2022hierarchical} introduce the pre-trained CLIP \cite{radford2021learning} model to T2I generation, demonstrating its efficacy in aligning images and text features. Rombach \etal \cite{rombach2022high} propose LDM, leveraging a powerful autoencoder to streamline the computational load of the iterative denoising process.
These pioneering efforts directly contribute to the inception of Stable Diffusion, elevating T2I generation to unprecedented levels of prominence within both the research community and the general public. 
Subsequent studies \cite{richardson2023conceptlab,sun2023sgdiff,kang2023counting,feng2022training,balaji2022ediffi, li2024unleashing} aim to improve the performance further. 
Notably, SD-XL \cite{podell2023sdxl} employs a larger backbone and incorporates diverse conditioning mechanisms, resulting in its ability to generate photo-realistic high-resolution images.
However, a notable limitation persists across these methods — they heavily rely on textual prompts as conditions, thus impeding precise control over the spatial composition of the generated image.

\subsection{Layout-to-image Synthesis}

Layout-to-image synthesis (LIS) revolves around generating images that conform to a prompt and corresponding layout instructions, \eg. bounding boxes or semantic masks. 
Several approaches \cite{yang2023reco,xue2023freestyle,zeng2023scenecomposer,li2023gligen,zhang2023adding,wang2023compositional,avrahami2023spatext,yang2023law,liu2023cones,wang2022semantic} suggest using paired layout-image data for training new models or fine-tuning existing ones.
For example, SceneComposer \cite{zeng2023scenecomposer} trains a layout-to-image model using a paired dataset of images and segmentation maps. In parallel, several approaches \cite{zhang2023adding, li2023gligen, wang2024instancediffusion, zhou2024migc, mou2023t2i} integrate additional components or adapters for layout control.
While these methods yield noteworthy results, they grapple with the challenge of labor-intensive and time-consuming data collection for training. Furthermore, a fully-supervised pipeline entails additional computational resource consumption and prolonged inference times.
 
Another series of methods \cite{he2023localized,kim2023dense,xie2023boxdiff,phung2023grounded,couairon2023zero,chen2023training,sun2024spatial} address the issue through a training-free approach with pre-trained models. Hertz \etal \cite{hertz2022prompt} initially observe that the spatial layouts of generated images are intrinsically connected with cross-attention maps. 
Building on this insight, Directed Diffusion\cite{ma2023directed} and DenseDiffusion \cite{kim2023dense} lead the way in manipulating the cross-attention map to align generated images with layouts. 
Subsequently, BoxNet \cite{wang2023compositional} propose a attention mask control strategy based on predicted object bounding boxes.
Some concurrent studies \cite{gong2023check, he2023localized} also propose various methods for modulating cross-attention maps.
Regrettably, even the state-of-the-art training-free approaches fall short in precise spatial control and suffer from semantic failures.

Closer to our work, several training-free approaches \cite{xie2023boxdiff, chen2023training, couairon2023zero, xiao2023r} design energy functions based on cross-attention maps to optimize the latent feature and encourage the desired objects to appear at the specified region.
However, our experiments revealed that these approaches lack precise spatial control as they rely solely on raw cross-attention maps extracted at each timestep, which are coarse-grained and noisy representations of desired objects.
Attention-Refocusing \cite{phung2023grounded} attempts to address this limitation by utilizing both cross-attention and self-attention maps individually for spatial control. However, it only optimizes the max values of attention maps, leading to unstable generation results and a lack of spatial accuracy.
In contrast, our $\mathcal{L}_{LAC}$ provides accurate guidance based on refined cross-attention maps, which are more precise representations of the desired objects. 
Therefore, $\mathcal{L}_{LAC}$ enhances the alignment between cross-attention maps and layout instructions and addresses semantic failures effectively.
Chen \etal \cite{chen2023training} notice a counter-intuitive phenomenon that padding tokens, \textit{i.e.}, start-of-text tokens (\texttt{[SoT]}) and end-of-text tokens (\texttt{[EoT]}), inherently carry rich semantic and layout information. However, this observation has not been thoroughly explored and utilized.
Our $\mathcal{L}_{PTC}$ efficiently harnesses the information embedded in padding tokens, further enhancing the consistency between object appearance and layout instructions.

\section{Method}
Our method guides the synthesis process based on self-attention maps and cross-attention maps extracted from T2I diffusion models.
Specifically, LoCo consists of three steps: (a) Attention Aggregation, (b) Localized Attention Constraint, and (c) Padding Tokens Constraint.
We provide a detailed presentation of these steps in the following sections.

\subsection{Preliminaries}

\textbf{Cross-attention maps.}\quad T2I diffusion models utilize cross-modal attention between text tokens and latent features in the noise predictor to condition the image synthesis.
Given a text prompt $\mathbf{y}$, a pre-trained CLIP \cite{radford2021learning} encoder is used to get the text tokens $\mathbf{e}=f_{\operatorname{CLIP}}(\mathbf{y})\in \mathbb{R}^{n \times d_{e}}$, \textit{i.e.}, text embedding features.
The \textit{query} $\mathbf{Q}_{z}$ and \textit{key} $\mathbf{K}_{e} $
are the projection of latent feature $\mathbf{z}_{t}$ and text tokens $\mathbf{e}$, respectively. At cross-attention layer $l$, the cross-attention maps $\mathbf{A}^{c,l}$ can be acquired as follows:
\begin{equation}
    \mathbf{A}^{c,l}=\operatorname{Softmax}\left(\frac{\mathbf{Q}_{z} \mathbf{K}_{e}^{\top}}{\sqrt{d}} \right)\in [0,1]^{hw \times n},
\end{equation}
where
$\mathbf{A}^{c,l}$ contains $n$ spatial attention maps $\mathbf{A}^{c,l}=\{\mathbf{A}^{c,l}_{0},\ldots,\mathbf{A}^{c,l}_{n-1}\}$. $\mathbf{A}^{c,l}_{i} \in [0,1]^{h \times w}$ corresponds to the \textit{i-th} text token $\mathbf{e}_{i}$.

Please note that, unlike previous methods, we preserve cross-attention maps of the start-of-text token (\textit{i.e.}, $\mathtt{[SoT]}$) and the end-of-text token (\textit{i.e.}, $\mathtt{[EoT]}$). 

\textbf{Self-attention maps.}\quad Self-attention maps capture the pairwise similarities among spatial positions within the latent features $\mathbf{z}_{t}$. At self-attention layer $l$, the self-attention map $\mathbf{A}^{s,l}$ is derived from \textit{query} $\mathbf{Q}_{z}$ and \textit{key} $\mathbf{K}_{z}$ from latent feature $\mathbf{z}_{t}$ as follows:
\begin{equation}
    \mathbf{A}^{s,l}=\operatorname{Softmax}\left(\frac{\mathbf{Q}_{z} \mathbf{K}_{z}^{\top}}{\sqrt{d}} \right)\in [0,1]^{hw \times hw}.
\end{equation}

\textbf{Problem setup.}\quad
\label{sec:problem_setup}
For clarity, we consider the input layout as $k$ bounding boxes $\mathcal{B}=\{\mathbf{b}_{1}, \ldots, \mathbf{b}_{k}\}$ and a text prompt $\mathbf{y}$ containing $k$ corresponding phrases $\mathcal{W}=\{\mathbf{w}_{1}, \ldots, \mathbf{w}_{k}\}$.
$\mathbf{b}_{i}$ indicates the user-provide location for the \textit{i-th} object and $\mathbf{w}_{i}$ describes the desired object in detail. Before applying the proposed constraints, 
we transform and resize each bounding box $\mathbf{b}_{i}$ to its corresponding binary mask $\operatorname{Mask}(\mathbf{b}_{i})$.  
\begin{figure}[t]
\centering
\includegraphics[width=\textwidth]{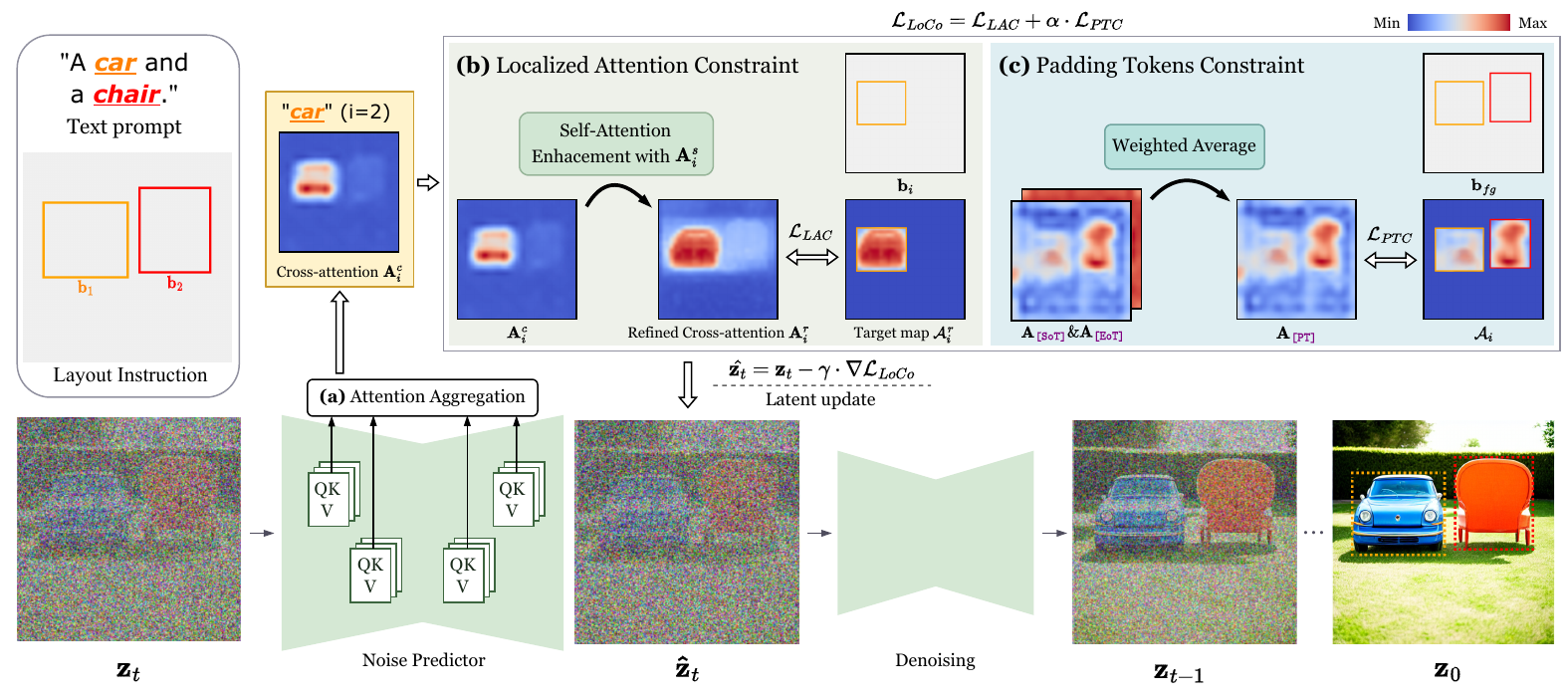}
\caption{
\textbf{Overview of LoCo.} 
LoCo consists of three steps: (a) Attention Aggregation, (b) Localized Attention Constraint, and (c) Padding Tokens Constraint.
At timestep $t$, we pass latent feature $\mathbf{z}_{t}$ through the noise predictor to extract cross-attention maps $\mathbf{A}^{c}$ and self-attention map $\mathbf{A}^{s}$. 
For the \textit{i-th} desired object, we obtain refined cross-attention map $\mathbf{A}^{r}_{i}$ via Self-Attention Enhancement to represent the object's appearance accurately. 
The proposed constraints, \textit{i.e.}, $\mathcal{L}_{LAC}$ and $\mathcal{L}_{PTC}$, are then applied to encourage the alignment between attention maps and layout instructions. Consequently, the latent feature $\mathbf{z}_{t}$ is updated with the $\triangledown\mathcal{L}_{LoCo}$ to obtain $\hat{\mathbf{z}_{t}}$ for denoising.
}
\label{fig:framework}
\end{figure}
\subsection{Attention Aggregation}
At each timestep $t$, the latent feature $\mathbf{z}_{t}$ is fed to the noise predictor of the T2I model. As shown in \cref{fig:framework} (a), we aggregate and average the attention maps across cross-attention layers and self-attention layers from the noise predictor respectively, obtaining aggregated attentions $\mathbf{A}^{c} \in [0,1]^{hw \times n}$ and $\mathbf{A}^{s} \in [0,1]^{hw \times hw}$:
\begin{equation}
    \begin{aligned}\mathbf{A}^{c}=\frac{1}{L}\sum_{l=1}^{L}\mathbf{A}^{c,l}\end{aligned}
    , \quad\begin{aligned}\mathbf{A}^{s}=\frac{1}{L}\sum_{l=1}^{L}\mathbf{A}^{s,l}.\end{aligned}
\end{equation}
\subsection{Localized Attention Constraint ($\mathcal{L}_{LAC}$)}
\label{sec:LAC}
\begin{figure}[t]
\centering
\includegraphics[width=\textwidth]{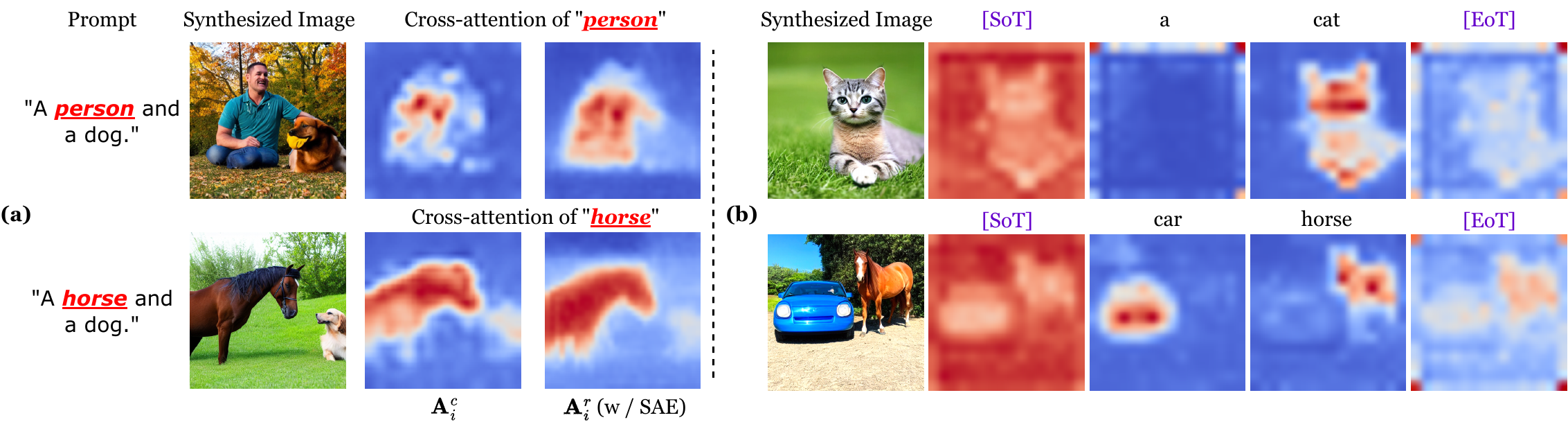}
\caption{
\textbf{(a) Visualization of Self-Attention Enhancement (SAE)}. SAE highlights the non-salient parts of the corresponding objects. Therefore, $\mathbf{A}^{r}_{i}$ serves as precise representations of desired objects.   \textbf{(b) Cross-attention maps of Padding Tokens}. One can observe from the examples that the padding tokens, i.e., start-of-text tokens ([SoT]) and end-of-text tokens ([EoT]) also carry substantial semantic and layout information.
}
\label{fig:sae_padding_token}
\end{figure}

Prior studies \cite{hertz2022prompt,xie2023boxdiff,kim2023dense} have demonstrated that the high-response regions in cross-attention maps perceptually align with synthesized objects in the decoded image. However, as shown in \cref{fig:sae_padding_token} (a), raw cross-attentions $\mathbf{A}^{c}_{i}$ only capture salient parts of the object and ignore non-salient ones, \eg, boundary regions. Hence, they are coarse-grained and noisy representations of desired objects, which are insufficient for precise spatial control.

Recent works on generating synthetic datasets \cite{ma2023diffusionseg, nguyen2024dataset} utilize self-attentions to improve the consistency between synthetic images and corresponding segmentation masks.
Inspired by these approaches, we perform Self-Attention Enhancement (SAE), improving raw cross-attention $\mathbf{A}^{c}_{i} \in [0,1]^{h \times w}$ for more accurate representations of desired object with self-attention $\mathbf{A}^{s} \in [0,1]^{hw \times hw}$:
\begin{equation}
    \mathbf{A}^{r}_{i} =  \mathbf{A}^{c}_{i} + \eta (\mathbf{A}^{s} \mathbf{A}^{c}_{i}- \mathbf{A}^{c}_{i}),
    \quad\mathbf{A}^{r}_{i}\in [0,1]^{h \times w},
\end{equation}
where $\eta$ controls the enhancement strength of self-attention. Intuitively, this operation leverages pairwise semantic affinity between pixels in $\mathbf{A}^{s}$, expanding the cross-attention map to positions with high semantic similarity and reinforcing non-salient regions (\cref{fig:sae_padding_token} (a)).
The refined cross-attention map $\mathbf{A}^{r}_{i}$ serves as improved description of the shape and position of the \textit{i-th} desired object.

Subsequently, we align $\mathbf{A}^{r}_{i}$ with its associated binary mask $\operatorname{Mask}(\mathbf{b}_{i})$ using $\mathcal{L}_{LAC}$ (\cref{fig:framework} (b)). We derive $\mathcal{A}_{i}$ by masking out elements of the cross-attention map beyond the target regions:
\begin{equation}
\mathcal{A}_{i} = \mathbf{A}^{r}_{i} \odot \operatorname{Mask}(\mathbf{b}_{i}),
\end{equation}
and the formulation of $\mathcal{L}_{LAC}$ is:
\begin{equation}
    \mathcal{L}_{LAC}=\sum\limits_{i=1}^{k}\left[ 1-\frac{ \sum_{x,y}(\frac{\mathcal{A}_{i} }{\| \mathbf{A}^{r}_{i} \|_\infty})}{ \sum_{x,y}(\frac{\mathbf{A}^{r}_{i}}{\| \mathbf{A}^{r}_{i} \|_\infty})} \right] ^{2},
\end{equation}
where $\sum_{x,y}$ means that we accumulate the value of each spatial entry in the cross-attention map. 
As shown in \cref{fig:vis_ablation}, $\mathcal{L}_{LAC}$ encourages high values to shift from the current high-activation regions into the corresponding target regions, guiding the \textit{i-th} desired object to appear at the specified location.

In contrast to energy functions proposed in previous methods\cite{chen2023training,phung2023grounded,xie2023boxdiff}, we normalize each refined cross-attention map $\mathbf{A}^{r}_{i}$ individually with $\| \mathbf{A}^{r}_{i} \|_\infty$.
This normalization is crucial because, although the high-response regions in the cross-attention map perceptually align with the positions of synthesized objects in the image, the maxima of these regions are numerically small (around 0.1) and fluctuating.
Normalization distinguishes high-response regions from the background, leading to accurate spatial control and preventing semantic inconsistencies.

\subsection{Padding Tokens Constraint ($\mathcal{L}_{PTC}$)}
$\mathcal{L}_{LAC}$ effectively encourages cross-attentions to focus on the correct regions. However, when these specified regions are located close together, the desired objects sometimes go beyond their corresponding boxes, causing misalignment between synthetic images and layout instructions.

To address this issue, we introduce the Padding Tokens Constraint (\cref{fig:framework} (c)).
As depicted in Fig. \ref{fig:sae_padding_token} (b), cross-attentions of both $\mathtt{[SoT]}$ and $\mathtt{[EoT]}$ tokens contain information about the image layout. While $\mathtt{[SoT]}$ primarily emphasizes the background, $\mathtt{[EoT]}$ responds to the foreground complementarily.
We leverage this semantic information in padding tokens to prevent objects from moving out of the target regions. 
Initially, we derive the mask for all foreground objects $\mathbf{b}_{fg}$:
\begin{equation}
    \operatorname{Mask}(\mathbf{b}_{fg})=\operatorname{Mask}(\bigcup_{i=1}^{k}\mathbf{b}_{i}),
\end{equation}
and obtain $\mathbf{A}_{\mathtt{PT}}$, cross-attention for padding tokens. $\mathbf{A}_{\mathtt{PT}}$ is a weighted average of the reversion of normalized $\mathbf{A}_{\mathtt{SoT}}$ and normalized $\mathbf{A}_{\mathtt{EoT}}$:
\begin{equation}
    \mathbf{A}_{\mathtt{PT}}=\beta\cdot\frac{1-\mathbf{A}_{\mathtt{SoT}}}{\| 1-\mathbf{A}_{\mathtt{SoT}} \|_\infty}+(1-\beta)\frac{\mathbf{A}_{\mathtt{EoT}}}{\| \mathbf{A}_{\mathtt{EoT}} \|_\infty},
\end{equation}
in which $\beta$ serves as a weighting factor. 

Subsequently, we define $\mathcal{L}_{PTC}$ as below:
\begin{equation}
    \mathcal{L}_{PTC}= \mathcal{L}_{\operatorname{BCE}}\left[\,\operatorname{Sigmoid} (\mathbf{A}_{\mathtt{PT}}), (\mathbf{A}_{\mathtt{PT}} \odot \operatorname{Mask}(\mathbf{b}_{fg}))\right].
\end{equation}
As shown in \cref{fig:vis_ablation}, $\mathcal{L}_{PTC}$ helps to penalize the erroneous activations that attend the background area, effectively preventing the incorrect expansion of desired objects.

\subsection{Latent Feature Update}
At each timestep $t$, the overall constraint $\mathcal{L}_{LoCo}$ is the weighted summation of $\mathcal{L}_{LAC}$ and $\mathcal{L}_{PTC}$ as follows:
\begin{equation}
    \mathcal{L}_{LoCo} = \mathcal{L}_{LAC}+\alpha\cdot\mathcal{L}_{PTC},
\end{equation}
where $\alpha$ is a factor controlling the intervention strength of $\mathcal{L}_{PTC}$.
We update the current latent feature $\mathbf{z}_t$ via backpropagation with $\mathcal{L}$ as below:
\begin{equation}
\label{eq:8}
    \hat{\mathbf{z}_t} \leftarrow  \mathbf{z}_t-\gamma\cdot\triangledown
\mathcal{L}_{LoCo}.
\end{equation}
Here, $\gamma$ is a scale factor controlling the strength of the guidance. Subsequently, $\hat{\mathbf{z}_t}$ is sent to the noise predictor for denoising.

Guided by $\mathcal{L}_{LoCo}$, $\mathbf{z}_{t}$ gradually adjusts at each timestep, aligning high-response attention regions to the specified bounding boxes. This process leads to the synthesis of desired objects in the user-provided locations.
Please refer to the experiments section for additional details.

\section{Experiments}
\label{sec:experiments}

 \begin{figure}[t]
\centering
\includegraphics[width=\textwidth]{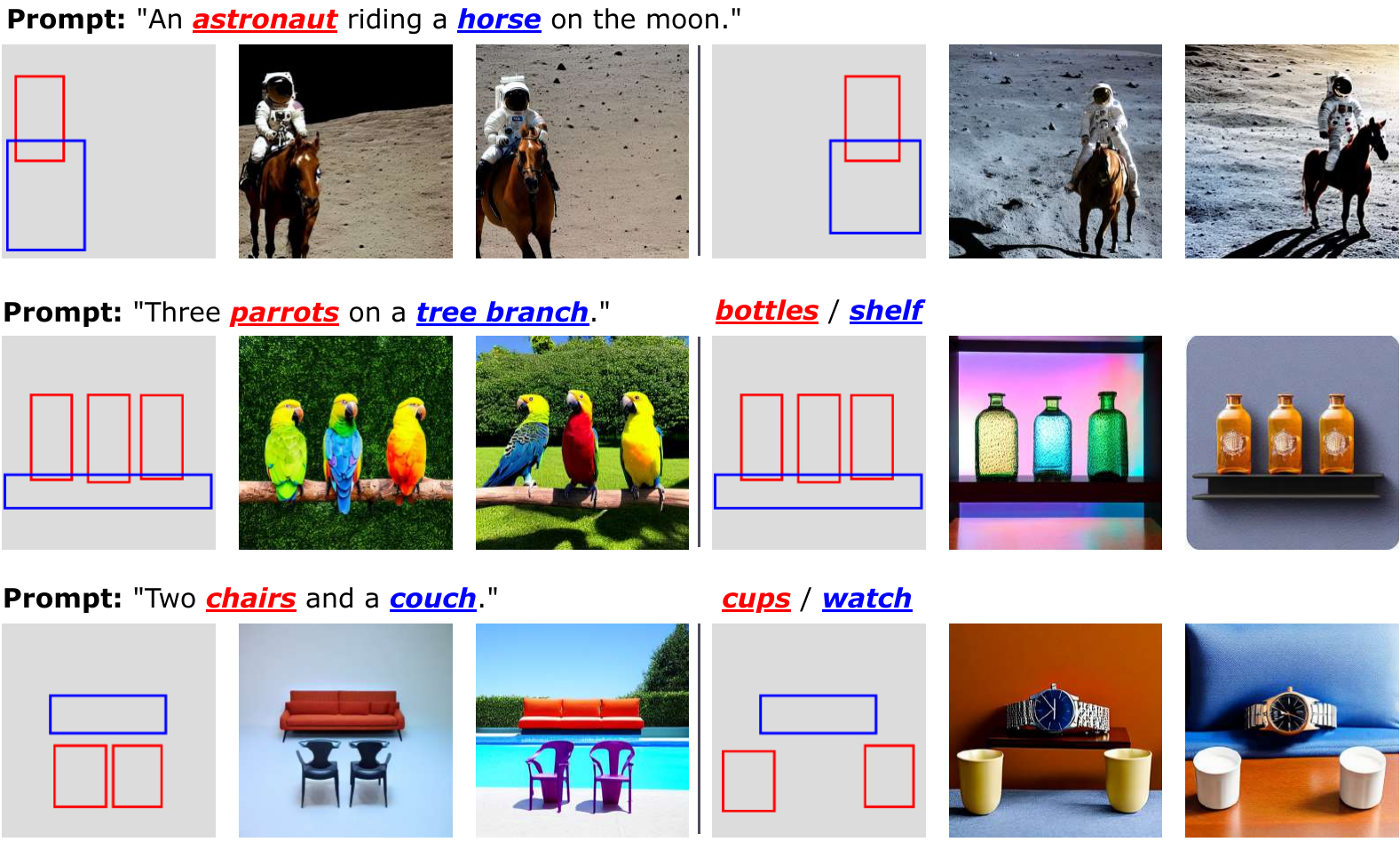}
\caption{
\textbf{Synthesized images with various conditioning inputs}, \textit{e.g.}, different locations and desired objects. LoCo is able to handle various spatial layouts and novel scenes while maintaining high image synthesis capability and precise concept coverage.
}
\label{fig:var_locations}
\end{figure}
\subsection{Experimental Setup}

\textbf{Datasets.} 
We conduct experiments on two standard benchmarks, the \textbf{HRS-Bench} \cite{bakr2023hrs} and the \textbf{DrawBench} \cite{saharia2022photorealistic}.
The \textbf{HRS-Bench} serves as a comprehensive benchmark for T2I models, offering various prompts divided into three main topics: accuracy, robustness, and generalization. 
As our method focuses on layout control, we specifically select four categories corresponding to image compositions from HRS: \textit{Spatial relationship}, \textit{Size}, \textit{Color} and \textit{Object Counting}. The number of prompts for each category is 1002/501/501/3000, respectively. The \textbf{DrawBench} dataset is a challenging benchmark for fine-grained analysis of T2I models. We utilize the category of \textit{Object counting} and \textit{Positional}, including 39 prompts.
Since both HRS and DrawBench do not include layout instructions, 
we incorporate publicly available layouts published by Phung et al. \cite{phung2023grounded} for evaluation.
To further evaluate our method’s capability in interpreting fine-grained layouts in the form of semantic masks, we utilize the dataset provided by \textbf{DenseDiffusion} \cite{kim2023dense}, which includes 250 binary masks with corresponding labels and captions.
To assess the performance of LoCo in synthesizing photorealistic images, we curate a COCO subset by randomly selecting 100 samples, along with their corresponding captions and bounding boxes from the \textbf{MS-COCO} \cite{lin2014microsoft} dataset.
\begin{figure}[t]
\centering
\includegraphics[width=\textwidth]{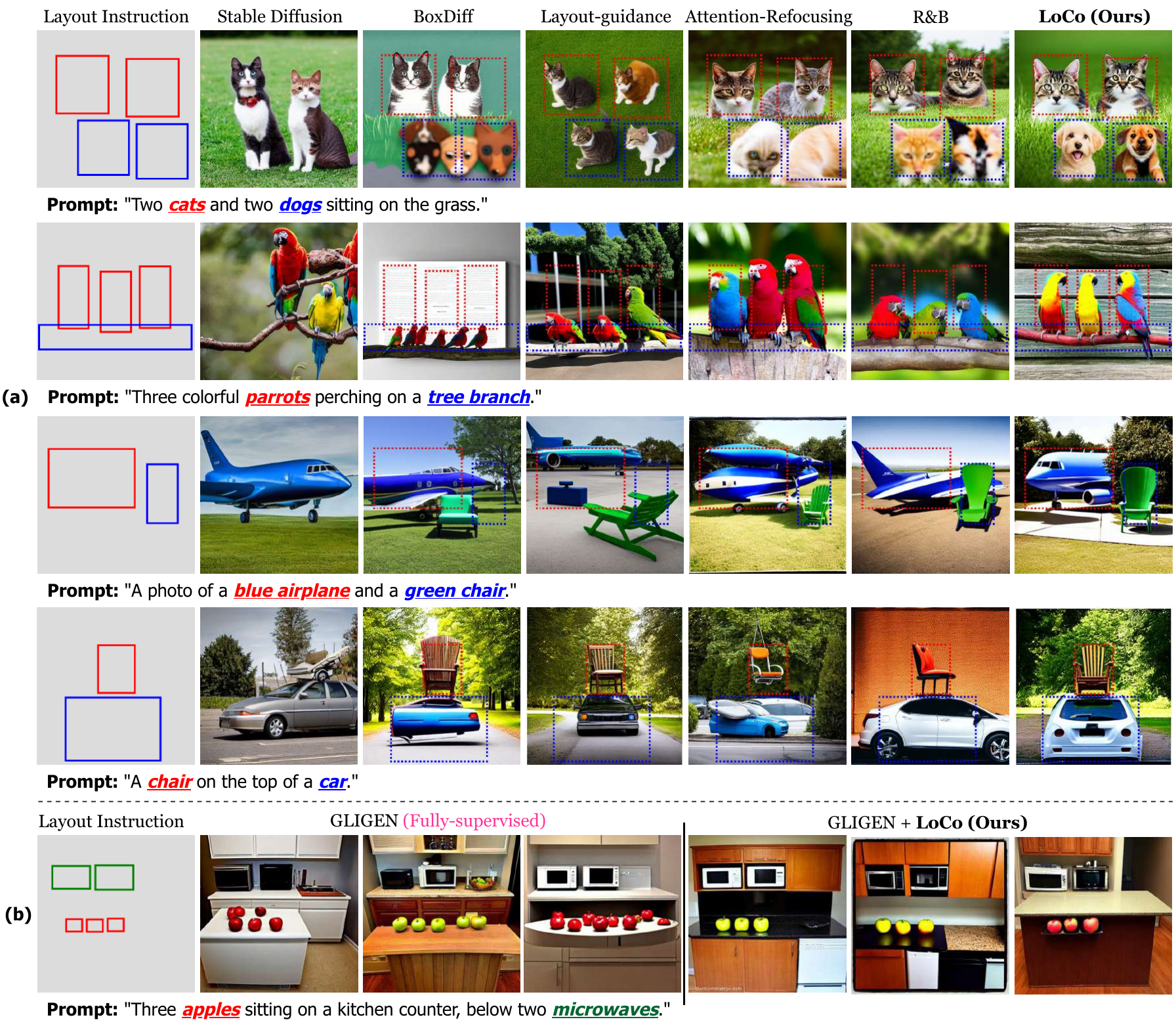}
\caption{
\textbf{(a) Visual comparisons with previous methods.} We show visual comparisons between LoCo and several training-free layout-to-image methods. The layout instructions are annotated on the images with dashed boxes. Our results faithfully adhere to both textual and layout conditions, outperforming prior approaches in terms of spatial control and image quality. \textbf{(b) Performance boost on fully-supervised layout-to-image method.} LoCo enhances the performance of GLIGEN \cite{li2023gligen} in generating multiple small objects significantly. Please zoom in for better view.
}
\label{fig:comparison}
    \vspace{-10pt}
\end{figure}

\textbf{Evaluation Metrics.} 
We follow the standard evaluation protocol of HRS. 
Specifically, we employ the pre-trained UniDet \cite{zhou2022simple}, a multi-dataset detector, on all synthesized images. 
Predicted bounding boxes are then utilized to validate whether the conditioning layout is grounded correctly.

For Spatial Compositions, \ie, the categories of \textit{Spatial relationship}, \textit{Size} and \textit{Color}, \textbf{generation accuracy} serves as the evaluation metric. A synthesized image is counted as a correct prediction when all detected objects, whether for spatial relationships, color, or size, are accurate. 
For \textit{Object Counting}, the number of objects
detected in generated images is compared to the ground
truths in text prompts to measure the \textbf{precision}, \textbf{recall}, and
\textbf{F1 score}. False positive samples happen when the number
of generated objects is smaller than the ground truths. In
contrast, the false negatives are counted for the missing objects.

For the DenseDiffusion \cite{kim2023dense} dataset and curated COCO subset, we report \textbf{IoU} and $\mathbf{AP}_{50}$ to measure the alignment of the input layout and synthesized images. Additionally, we employ \textbf{CLIP score} to evaluate the fidelity of synthesized images to textual conditions.

\begin{figure}[t]
\centering
\includegraphics[width=\textwidth]{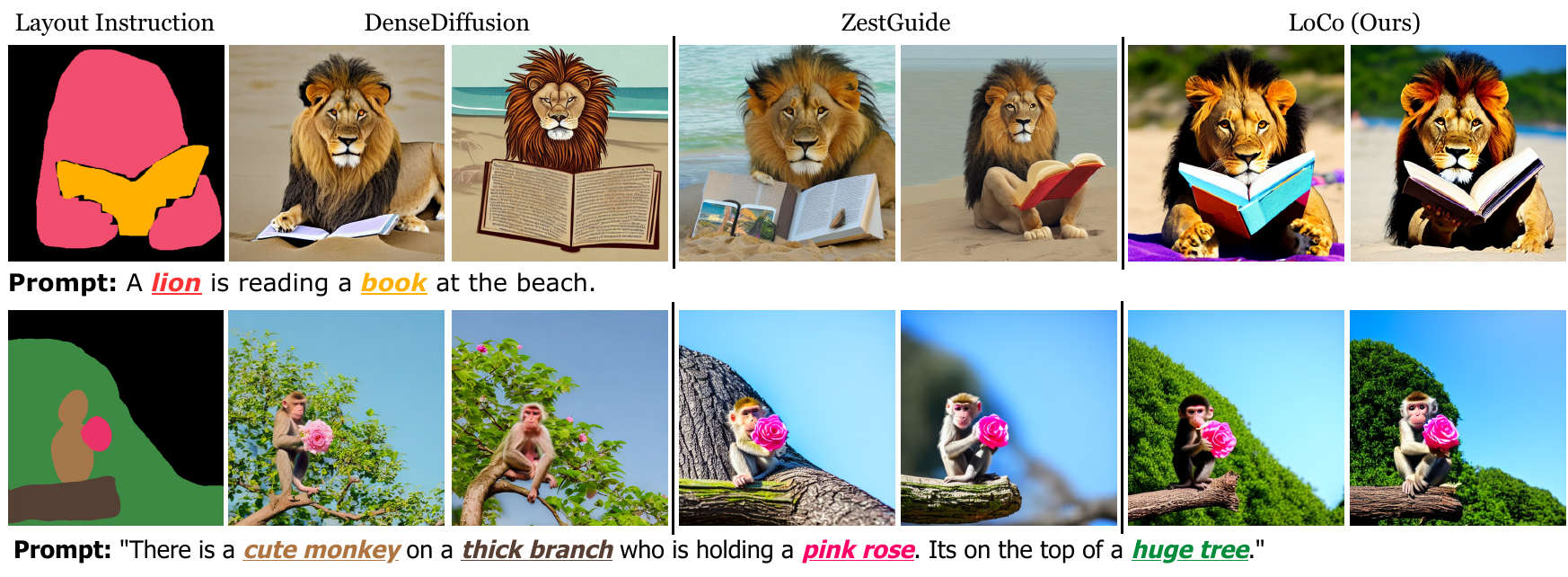}
\caption{
\textbf{Visual comparisons with training-free layout-to-image methods on mask-level layout
instructions.} Our results faithfully adhere to the fine-grained layout conditions. 
}

\label{fig:mask_comparison}
\end{figure}

\textbf{Implementation Details.} 
\label{sec:implementation}
Unless specified otherwise, we use the official Stable Diffusion V-1.4 \cite{rombach2022high} as the base T2I synthesis model. 
The synthesized images, with a resolution of $512\times512$, are generated with 50 denoising steps. 
For the hyperparameters, we use the loss scale factor $\gamma = 30$, $\eta=0.3$ for Self-Attention Enhancement, $\alpha=0.2$ and $\beta = 0.8$. Classifier-free guidance \cite{ho2022classifier} is utilized with a fixed guidance scale of 7.5.
Given that the layout of the synthesized image is typically established in early timesteps of inference, we integrate guidance with proposed constraints during the initial 10 steps. In each timestep, the latent update in Eq. (\ref{eq:8}) iterates 5 times before denoising.

\subsection{Qualitative Results}
\textbf{Visual variations.} 
As depicted in \cref{fig:var_locations}, to validate the robustness of our proposed method, we vary the textual prompts and layout instructions to synthesize different images. 
In the $1^{st}$ row, we shift the location of "astronaut" and "horse" from the left to the right of the image. 
LoCo produces delicate results in accordance with the instructions.
In the $2^{nd}$ and $3^{rd}$ rows, we change the layout instructions and desired objects simultaneously.
The synthetic images follow the user-provided conditions faithfully across multiple spatial locations and prompt variations.
This demonstrates that our method can handle various spatial layouts and textual prompt while maintaining high image synthesis fidelity and precise concept coverage. 

\textbf{Comparisons with prior methods.} 
Fig. \ref{fig:comparison} (a) provides a visual comparison of various state-of-the-art training-free LIS methods, illustrating that our proposed LoCo consistently facilitates the synthesis of images which faithfully adhere to the layout conditions.
For instance, as shown in the $1^{st}$ row, given a prompt like ``Two cats and two dogs sitting on the grass.'' and a layout instruction, BoxDiff \cite{xie2023boxdiff} fails in placing the dogs.
Attention-Refocusing \cite{phung2023grounded} and Layout-guidance \cite{chen2023training} correctly generate two cats according to their respective boxes, but they suffer from missing or fused dogs. 
R\&B \cite{xiao2023r} exhibits good spatial controllability but generates four cats erroneously. 
In contrast, LoCo accurately generates both the ``cat'' and ``dog'' based on the given layout.
In the $2^{nd}$ and $4^{th}$ rows, LoCo faithfully positions the desired objects according to the conditioning layout, while competing methods suffer from inaccurate spatial control.

Moreover, we observe that when integrated into a fully-supervised layout-to-image method, such as GLIGEN \cite{li2023gligen}, LoCo significantly improves GLIGEN's performance in generating multiple small objects (\cref{fig:comparison} (b)).

\begin{table}[t]
    \centering
    \caption{Comparison with training-free layout-to-image synthesis methods on \textbf{image compositions}. We report the inference time of these methods on a single NVIDIA RTX 3090 GPU. $^{\dagger}$: A\&R denotes Attention-Refocusing \cite{phung2023grounded}.}
    \resizebox{\linewidth}{!}{
    \renewcommand{\arraystretch}{1}
        \begin{tabular}{l l c c c c c c}
            \toprule[1.2pt]
            \multirow{2.5}{*}{Method} & \multirow{2.5}{*}{Venue} & \multicolumn{3}{c}{HRS-Bench} & \multicolumn{1}{c}{DrawBench} & \multirow{2.5}{*}{CLIP($\uparrow$)} & {Inference }   \\
            \cmidrule(lr){3-5} \cmidrule(lr){6-6}  
            & &  Spatial$(\uparrow)$ & Size$(\uparrow)$ & Color$(\uparrow)$ &  Positional$(\uparrow)$ & &Time($\downarrow$) \\
            \midrule
            Stable Diffusion\cite{rombach2022high} & CVPR'22 & 10.08 & 12.05 & 13.01 & 12.50 & 0.3070 & 8.15 s \\
            Attend-and-Excite\cite{chefer2023attend} & SIGGRAPH'23 & 14.15 & 13.28 & 18.23 & 20.50 & 0.3081 & 25.43 s\\
            MultiDiffusion\cite{bar2023multidiffusion}  &  ICML'23  & 16.86 & 13.54 & 17.55 & 36.00 & 0.3096 & 19.57 s\\

            DenseDiffusion\cite{kim2023dense} &  ICCV'23   & 17.56 & 14.31 & 18.27 & 30.50 & 0.3094 & 11.54 s\\
            BoxDiff\cite{xie2023boxdiff} & ICCV'23 & 16.52 & 13.35 & 14.51 & 32.50 & 0.3125 & 32.50 s\\ 

            Layout-guidance\cite{chen2023training} &  WACV'24   & 22.06 & 15.83 & 15.36 & 36.50 & 0.3148 & 22.75 s \\
            
            A\&R$^{\dagger}$\cite{phung2023grounded} & CVPR'24  & 24.55 & 16.63 & 21.31 & 43.50 & 0.3140 & 55.49 s\\
            R\&B\cite{xiao2023r}     &  ICLR'24   & 29.57 & 25.51 & 30.27  & 55.00 & 0.3167 & 34.32 s \\
            LoCo (Ours)     &  -   & \textbf{34.86} & \textbf{26.50} & \textbf{31.50}  & \textbf{55.50} & \textbf{0.3179} & 20.22 s \\

            \bottomrule[1.2pt] 
    \end{tabular}
    }
    \label{tab:main_comp}
\end{table}
\begin{table}[ht]
    \centering
    \caption{Comparison with training-free layout-to-image synthesis methods on \textbf{object counting}. $^{\dagger}$: A\&R denotes Attention-Refocusing \cite{phung2023grounded}.}
    \resizebox{\linewidth}{!}{
    \renewcommand{\arraystretch}{1}
        \begin{tabular}{l l c c c c c c c }
            \toprule[1.2pt]
            \multirow{2.5}{*}{Method} & \multirow{2.5}{*}{Venue} & \multicolumn{3}{c}{HRS-Bench} & \multicolumn{3}{c}{DrawBench} & \multirow{2.5}{*}{CLIP($\uparrow$)}   \\
            \cmidrule(lr){3-5} \cmidrule(lr){6-8}  
            & &  Precision$(\uparrow)$ & Recall$(\uparrow)$ & F1$(\uparrow)$ &  Precision$(\uparrow)$ & Recall$(\uparrow)$ & F1$(\uparrow)$ & \\
            \midrule
            Stable Diffusion\cite{rombach2022high} & CVPR'22 & 71.86 & 52.19 & 58.31 & 73.32& 70.00& 71.55 & 0.3081  \\
            Attend-and-Excite\cite{chefer2023attend} & SIGGRAPH'23 & 73.10 & 54.79 & 60.47 & 77.64 & 74.85 & 76.20 & 0.3079\\
            MultiDiffusion\cite{bar2023multidiffusion}  &  ICML'23  & 80.60 & 45.83 & 56.22 & 75.37 & 65.61 & 69.90 & 0.3099 \\

            DenseDiffusion\cite{kim2023dense} &  ICCV'23   & 82.21 & 51.32 & 63.19  & 78.46 & 72.54 & 75.38 & 0.3113\\
            BoxDiff\cite{xie2023boxdiff} & ICCV'23 & 81.54 & \textbf{56.61} & 66.83  & 75.16 & 71.55 & 73.28 & 0.3126 \\

            Layout-guidance\cite{chen2023training} &  WACV'24   & 80.60 & 45.83 & 56.22 & 79.15 & 70.61 & 74.48 & 0.3124 \\
            
            A\&R$^{\dagger}$\cite{phung2023grounded} & CVPR'24  & 81.56 & 51.19 & 60.62 & 78.53& 73.63 & 75.81 & 0.3143\\
            R\&B\cite{xiao2023r}     &  ICLR'24   &  83.35 & 56.08  & 67.04   & 83.74 & \textbf{82.89} & 83.31 & 0.3152 \\
            LoCo (Ours)     & -    & \textbf{84.91} & 55.52 & \textbf{67.14} & \textbf{89.69} & 81.15 & \textbf{85.21} & \textbf{0.3158} \\

            \bottomrule[1.2pt] 
    \end{tabular}
    }
    \label{tab:counting_comp}
    \vspace{-10pt}
\end{table}
\subsection{Quantitative Results}
\textbf{Box-level Layout Instruction.}\quad We compare LoCo with various state-of-the-art training-free LIS methods based on the Stable Diffusion V-1.4 in Table. \ref{tab:main_comp} and Table. \ref{tab:counting_comp}.

Our approach demonstrates remarkable accuracies across all categories on the HRS-Bench compared to prior layout-to-image methods. In the DrawBench, LoCo also delivers a noteworthy performance improvement over the standard Stable Diffusion, showcasing its proficiency in interpreting fine-grained spatial conditions. 
This enhancement can be attributed to that LoCo effectively reinforces the alignment between object appearance and layout instructions with precise spatial control. Furthermore, LoCo outperforms previous approaches in image quality, as evidenced by higher CLIP scores, suggesting that our approach achieves superior alignment between synthesized images and textual prompts. Moreover, our proposed LoCo achieves a good balance between inference time and spatial controllability.

The integration of LoCo also significantly boosts the performance of GLIGEN \cite{li2023gligen}, as depicted in Table. \ref{tab:gligen_comp}. 
This underscores the versatility of LoCo, serving as a plug-and-play booster for fully-supervised layout-to-image methods. 

\textbf{Mask-level Layout Instruction.}\quad LoCo also smoothly extends to various forms of layout instructions, \eg, semantic masks (\cref{tab:mask_comp}, \cref{fig:mask_comparison}). Our method outperforms the current state-of-the-art approaches with higher mIoU and CLIP score, indicating LoCo's superiority in fine-grained spatial control.

\begin{table}[t]
    \centering
    \caption{LoCo serves as a plug-and-play booster when integrated into fully-supervised layout-to-image method, \textit{e.g.}, GLIGEN \cite{li2023gligen}.}
    \resizebox{\linewidth}{!}{
    \renewcommand{\arraystretch}{1}
        \begin{tabular}{l l c c c c c c c }
            \toprule[1.2pt]
            \multirow{2.5}{*}{Method} & \multirow{2.5}{*}{Venue} & \multicolumn{4}{c}{HRS-Bench} & \multicolumn{2}{c}{DrawBench} & \multirow{2.5}{*}{CLIP($\uparrow$)}   \\
            \cmidrule(lr){3-6} \cmidrule(lr){7-8}  
            & &  Spatial$(\uparrow)$ & Size$(\uparrow)$ & Color$(\uparrow)$ & Counting(F1)&  Positional$(\uparrow)$ &  Counting(F1) &\\
            \midrule
            GLIGEN\cite{li2023gligen}     & CVPR'23    & 40.22 & 32.13 & 16.17 & 68.32 & 46.50 & 81.68 & 0.3167\\
             + A\&R\cite{phung2023grounded}    &      & 53.69 & 39.96 & 23.71 & 71.83 & 64.50 & 87.61 & 0.3198\\
             + R\&B\cite{xiao2023r}   &     & 56.87 & 42.69 & \textbf{35.72} & 74.57 & 67.50& 88.58 & 0.3232\\
             + LoCo (Ours)   &     & \textbf{59.48} & \textbf{43.37} & 35.45 & \textbf{76.24} & \textbf{72.00} & \textbf{89.26} & \textbf{0.3242}\\
            \bottomrule[1.2pt] 
    \end{tabular}
    }
    \label{tab:gligen_comp}
\end{table}
\begin{table}[ht]
    \scriptsize
    \centering
    \tabcolsep=0.4cm
    \caption{Comparison with training-free layout-to-image methods on mask-level layout instructions.}
    \renewcommand{\arraystretch}{1}
        \begin{tabular}{l l c c }
            \toprule[1pt]
            Method & Venue & \quad mIoU($\uparrow$)  & CLIP($\uparrow$)   \\

            \midrule
            SD-Pww\cite{balaji2022ediffi}     & arXiv'22    & 23.76 
            $\pm$ 0.50 & 0.2800 $\pm$ 0.0005  \\
            DenseDiffusion\cite{kim2023dense}     & ICCV'23     & 34.99 $\pm$ 1.13 & 0.2814 $\pm$ 0.0005  \\
            ZestGuide\cite{couairon2023zero}     & ICCV'23     & 40.15 $\pm$ 0.24 & 0.3174 $\pm$ 0.0008  \\            
            A\&R\cite{phung2023grounded}     & CVPR'24     & 38.97 $\pm$ 0.56 & 0.3177 $\pm$ 0.0011 \\
            LoCo (Ours)     &   -  & \textbf{43.12 $\pm$ 0.62} & \textbf{0.3188 $\pm$ 0.0016}  \\
            \bottomrule[1pt] 
    \end{tabular}
    \label{tab:mask_comp}
\end{table}
\subsection{Ablation Studies}

\begin{figure}[t]
\centering
\includegraphics[width=0.9\textwidth]{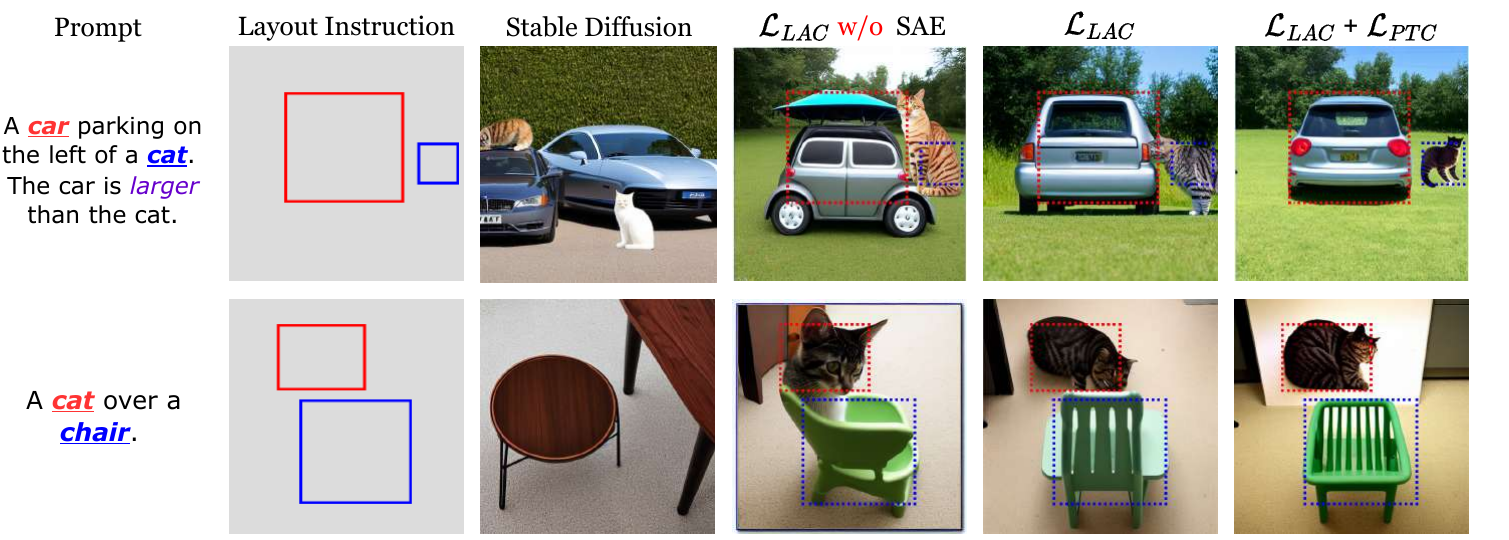}
\caption{
\textbf{Visual Ablations: Impact of Different Components of LoCo.} The layout instructions are annotated on the images with dashed boxes. 
}
\label{fig:vis_ablation}
\end{figure}

\textbf{Ablation of Key Components.}
We investigate the effectiveness of critical components in our method on the DenseDiffusion dataset and HRS-Bench,  as outlined in Table. \ref{tab:ablation_comp}. Visualized results are provided in Fig. \ref{fig:vis_ablation}.

We first assess the impact of $\mathcal{L}_{LAC}$. $\mathcal{L}_{LAC}$ exhibits effectiveness even without SAE and correctly interprets spatial relationships of the desired objects. This suggests that $\mathcal{L}_{LAC}$ without SAE is inherently advanced in controlling the spatial composition of synthesized images. However, $\mathcal{L}_{LAC}$ without SAE does not provide accurate spatial control (see the $4^\mathbf{th}$ column of \cref{fig:vis_ablation}). Introducing SAE in $\mathcal{L}_{LAC}$ results in a substantial performance boost in mIoU and improves the consistency between synthetic images and corresponding layout instructions (see the $5^\mathbf{th}$ column of \cref{fig:vis_ablation}). 

Moreover, solely employing $\mathcal{L}_{PTC}$ provides a degree of spatial control compared to vanilla Stable Diffusion. This underscores that padding tokens also carry substantial semantic and layout information.
\begin{table}[t]
  \centering
    \caption{Ablation study on LoCo's key components and impact of hyper-parameters.}
        \vspace{-10pt}
  \begin{subtable}{0.68\linewidth}
    \caption{Ablations on various combinations of components. We report performance on the DenseDiffusion dataset and HRS-Bench.}
    \resizebox{\linewidth}{!}{
    \renewcommand{\arraystretch}{1}
        \begin{tabular}{c c c c c c c c }
            \toprule[1.5pt]
            \multirow{2.5}{*}{$\mathcal{L}_{LAC}$ w/o SAE} &\multirow{2.5}{*}{$\mathcal{L}_{LAC}$} & \multirow{2.5}{*}{$\mathcal{L}_{PTC}$} &  \multicolumn{1}{c}{DenseDiffusion} & \multicolumn{3}{c}{HRS-Bench}  & \multirow{2.5}{*}{CLIP($\uparrow$)} \\
            \cmidrule(lr){4-4}  \cmidrule(lr){5-7}
            & & &  mIoU$(\uparrow)$ & Spatial$(\uparrow)$ & Size$(\uparrow)$ & Color$(\uparrow)$ & \\
            \midrule
            $\times$ & $\times$ & $\times$ & 9.15 &  10.08 & 12.05 & 13.01   &  0.3073 \\
            \checkmark &  $\times$ &  $\times$ & 34.55 & 29.21 & 22.64  & 27.86 &  0.3147   \\
             $\times$     & \checkmark & $\times$  &40.13  & 32.24 & 25.52 & 29.94 &  0.3159\\

            $\times$  & $\times$   & \checkmark   & 14.33 & 14.86 & 16.44  & 15.63 &  0.3096 \\

            \checkmark &  $\times$  & \checkmark & 36.17 & 31.54 & 24.05  & 28.92 &  0.3154 \\
       $\times$ & \checkmark& \checkmark & \textbf{43.12}  &   \textbf{34.86} & \textbf{26.50} & \textbf{31.50}  & \textbf{0.3161} \\

            \bottomrule[1.5pt] 
    \end{tabular}
    }
    \label{tab:ablation_comp}
  \end{subtable}
      \hfill
  \begin{subtable}{0.3\linewidth}
    \caption{Ablation study on loss scale $\gamma$.}
    \resizebox{\linewidth}{!}{
    \renewcommand{\arraystretch}{1}
        \begin{tabular}{c c c  }
            \toprule[1.3pt]

           Loss scale ($\gamma$)  & AP$_{50}$($\uparrow$) & CLIP($\uparrow$)  \\
           \midrule
           5 & 20.15 & 0.3054 \\
           10 & 27.85 & 0.3088 \\
           20 & 35.52 & 0.3108 \\
           30 & 51.54 & 0.3096 \\
           40 & 46.62 & 0.3087 \\
           50 & 39.79 & 0.3066 \\
           75 & 29.08 & 0.3010 \\

            \bottomrule[1.3pt] 
    \end{tabular}
    }
    \label{tab:loss_scale}
  \end{subtable}

    \vspace{-10pt}
  \label{tab:ablation}
\end{table}

Simultaneously utilizing $\mathcal{L}_{LAC}$ and $\mathcal{L}_{PTC}$ yields the best results in spatial controllability, considering both quantitative metrics and qualitative assessments (see the $7^\mathbf{th}$ column of \cref{fig:vis_ablation}). The synthetic images now faithfully adhere to both textual and layout conditions.

\textbf{Ablation on Loss Scale.} 
In Table. \ref{tab:loss_scale}, we explore the trade-off between spatial controllability and image fidelity by varying loss scale $\gamma$ from 5 to 75.
We report the AP$_{50}$ and CLIP score on the curated COCO subset.
Notably, as $\gamma$ grows, both scores initially improve before experiencing a rapid decline.  
This phenomenon signifies that excessively strong constraints significantly compromise generative fidelity, leading to a degradation in evaluation results.

Please refer to the supplementary for additional results and ablations.
\section{Conclusion}
This paper proposes LoCo, a training-free approach for layout-to-image synthesis. We introduce two novel constraints, \textit{i.e.}, $\mathcal{L}_{LAC}$ and $\mathcal{L}_{PTC}$, which excels in providing accurate spatial control and mitigating semantic failures faced by previous methods.
LoCo seamlessly integrates into existing text-to-image and layout-to-image models, amplifying their performance without the necessity for additional training or paired layout-image data. 
Extensive experiments showcase that LoCo significantly outperforms existing training-free layout-to-image approaches by a substantial margin.

%
%
\bibliographystyle{splncs04}
\bibliography{egbib}
\end{document}